\documentclass[conference]{IEEEtran}
\IEEEoverridecommandlockouts
\usepackage{cite}
\usepackage{amsmath,amssymb,amsfonts}
\usepackage{algorithmic}
\usepackage{graphicx}
\usepackage{textcomp}
\usepackage{xcolor}

\usepackage{balance}
\usepackage{makecell}
\usepackage{booktabs}
\usepackage{multirow}
\usepackage{array}
\usepackage[most]{tcolorbox}
\usepackage{soul}

\usepackage[caption=false,font=footnotesize]{subfig}

\usepackage[pagebackref,breaklinks,colorlinks,allcolors=cvprblue]{hyperref}
\usepackage{flushend}
\usepackage{placeins}
\usepackage{tabularx}
\usepackage{enumitem}
\usepackage{fontawesome5}

\def\BibTeX{{\rm B\kern-.05em{\sc i\kern-.025em b}\kern-.08em
    T\kern-.1667em\lower.7ex\hbox{E}\kern-.125emX}}

\definecolor{cvprblue}{rgb}{0.21,0.49,0.74}
\definecolor{marker}{RGB}{190,220,255} % light blue
\sethlcolor{red!50}

\newtcolorbox{takeawaybox}{
  colback=blue!8,        % light blue background
  colframe=blue!60!black,
  arc=6pt,               % rounded corners
  boxrule=1pt,
  left=2pt,
  right=2pt,
  top=6pt,
  bottom=6pt
}

\newtcolorbox{promptbox}{
  colback=blue!3,        % light blue background
  colframe=blue!60!black,
  arc=6pt,               % rounded corners
  boxrule=1pt,
  left=2pt,
  right=2pt,
  top=2pt,
  bottom=2pt
}

\begin{document}

\title{MLLM-based Textual Explanations for Face Comparison}

\author{
\IEEEauthorblockN{
Redwan Sony \quad Anil K. Jain \quad Arun Ross
}
\IEEEauthorblockA{
Department of Computer Science and Engineering, Michigan State University, East Lansing, MI, USA\\
\texttt{\{sonymd, jain, rossarun\}@msu.edu}
}
}

\maketitle

\begin{abstract}
Multimodal Large Language Models (MLLMs) have recently been proposed as a means to generate natural-language explanations for face recognition decisions. While such explanations facilitate human interpretability, their reliability on unconstrained face images remains underexplored. In this work, we systematically analyze MLLM-generated explanations for the unconstrained face verification task on the challenging IJB-S dataset, with a particular focus on extreme pose variation and surveillance imagery. Our results show that even when MLLMs produce correct verification decisions, the accompanying explanations frequently rely on non-verifiable or hallucinated facial attributes that are not supported by visual evidence. We further study the effect of incorporating  information from traditional face recognition systems, viz., similarity scores and match decisions provided alongside the input images. Although such information improves categorical verification performance, it does not consistently lead to faithful explanations. To evaluate the explanations beyond decision accuracy, we introduce a likelihood-ratio–based framework that measures the evidential strength of textual explanations. Our findings highlight fundamental limitations of current MLLMs for explainable face recognition and underscore the need for a principled evaluation of reliable and trustworthy explanations in biometric applications. Code is available  at \url{https://github.com/redwankarimsony/LR-MLLMFR-Explainability}.
\end{abstract}

\begin{IEEEkeywords}
Face recognition, multimodal large language models, explainable AI, likelihood ratio. 
\end{IEEEkeywords}

\begin{figure}[t]
\centering

\resizebox{0.95\linewidth}{!}{%
\begin{minipage}{\linewidth}
\begin{takeawaybox}
\textbf{Prompt:} Compare the following two face images and generate a detailed explanation of their similarities and differences following the required format. 

\centering
\includegraphics[width=0.6\linewidth]{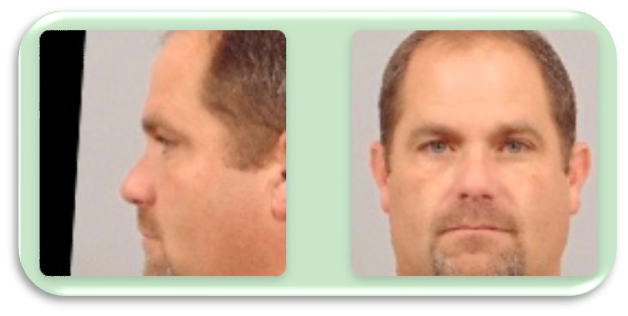}
\vspace{2pt}

\raggedright
\textbf{Match Verdict:} Match 

\textbf{Similarities:} Similar facial structure, including a broad forehead, \hl{a prominent jawline}, and \hl{similar nose morphology}. The skin tone appears consistent. The hairline and \hl{ear structure} are also alike. The periocular region, including the \hl{eyebrows and eye shape}, displays noticeable consistency. 

\textbf{Differences:} Image 1 is a profile view, while Image 2 presents a frontal view, \hl{which affects the visibility of certain features like the nose} and lips. Lighting conditions also differ slightly. The expression might vary slightly due to the angle, but the core features remain consistent.
\end{takeawaybox}
\end{minipage}
}%

\vspace{0.4em}
\caption{Example of GPT-4o~\cite{openai2024gpt4o} explanation for face verification. Although the model predicts the correct match decision, the highlighted text shows that the explanation relies on non-verifiable attributes, indicating weak visual grounding.}
\label{fig:visual_abstract}

\end{figure}

\section{Introduction}
\label{sec:intro}
Face recognition (FR) is the task of automatically identifying or verifying a person’s identity from facial imagery~\cite{minchul50years}. The use of deep neural network models has changed the landscape of FR systems, with improved accuracy even on unconstrained ``in-the-wild'' images~\cite{kprpe_2022_kim}. At the same time, recent advances in multimodal large language models (MLLMs) have enabled the generation of natural-language explanations for computer vision tasks~\cite{openai2023gpt4,deandres2401chatgpt,hassanpour2024chatgpt,shahreza2025facellm,li2025faceinsight,sony2025benchmarking}, including face recognition. By conditioning on images, similarity scores, or auxiliary metadata, MLLMs can produce human-readable explanations describing facial similarities and differences~\cite{sony2025foundation}, offering a promising pathway toward explainable face recognition. However, prior work has shown that such explanations may rely on linguistic priors rather than visual evidence, leading to hallucinated or weakly grounded reasoning~\cite{liu2024survey,bai2024hallucination,dang2025survey,huang2025survey,zhang2025siren,song2026large}. This raises critical concerns about the reliability of MLLM-based explanations, particularly in forensic and security applications where explanations may be interpreted as evidence.

In this work, we show that this issue is especially consequential for face recognition. Our experiments demonstrate that even when MLLMs produce the correct verification decision, the accompanying explanations may be inaccurate, unverifiable, or misleading, as shown in Fig.~\ref{fig:visual_abstract}, which undermines the reliability of MLLMs as explainability tools. An important question is whether auxiliary information from classical FR systems, such as similarity scores or match decisions, improves explanation reliability. To investigate this, we evaluate multiple MLLMs viz., GPT-4o~\cite{openai2024gpt4o} and Gemini-2.5~\cite{geminiteam2025geminifamilyhighlycapable}, augmented with outputs from domain-specific FR models on the challenging IJB-S Still-to-Still benchmark\cite{kalka2018ijbs}. While auxiliary FR information improves categorical verification performance, it does not consistently improve explanation faithfulness. In contrast, commercial off-the-shelf (COTS) FR systems achieve near-perfect verification accuracy but provide no explanations, highlighting a fundamental trade-off between accuracy and transparency.

To address this gap, we introduce a likelihood ratio (LR)-based framework that quantifies the evidential strength of textual explanations by MLLMs, independently of the correctness of the decision. By modeling the distributions of explanation embeddings under genuine and impostor hypotheses, our framework enables principled evaluation of explanation reliability using established evidence likelihood theory. Our analysis shows that hallucinated or weakly grounded explanations remain prevalent under unconstrained conditions, even when the verification accuracy of the MLLM improves. Our contributions are summarized as follows:

\begin{enumerate}[leftmargin=*]
    \item We systematically evaluate MLLM-generated textual explanations for face verification under extreme pose variation, showing a gap between decision correctness and explanation faithfulness.
    \item We analyze the effect of FR information (similarity scores and decisions) on MLLM verification performance.
    \item We introduce a likelihood-ratio–based framework to quantify the evidential strength of textual explanations beyond categorical (i.e., verification) accuracy.
    \item We provide empirical insights into when MLLMs produce visually grounded explanations versus relying on linguistic priors.
\end{enumerate}

\begin{figure*}[t]
    \centering
    \includegraphics[width=0.98\linewidth]{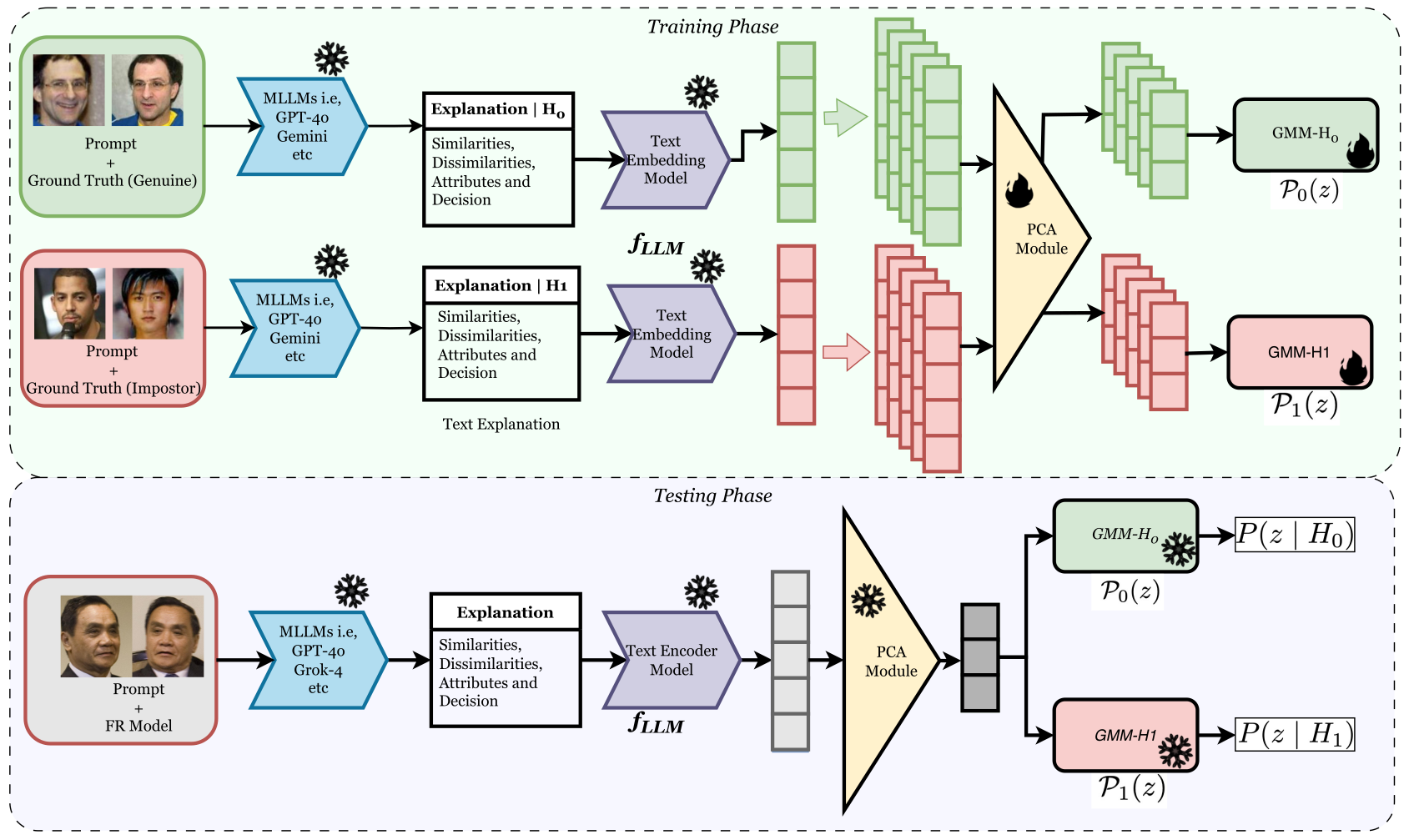}
    \vspace{-4pt}
\caption{Training and testing pipeline for LR-based evaluation of MLLM explanations. During training (top), explanations generated with ground-truth labels are embedded and used to learn class-conditional GMMs for genuine ($H_0$) and impostor ($H_1$) hypotheses; trainable modules are marked \faFire, and frozen modules are marked \faSnowflake. During testing (bottom), explanations generated under different prompting regimes, with or without auxiliary FR information, are embedded using the frozen pipeline and evaluated via text-derived likelihood ratios.}
 \label{fig:method-diagram}
\end{figure*}

\section{Literature Review}
Face recognition explanations generally fall into two categories: saliency map–based methods and textual explanations. Saliency-based approaches highlight image regions that contribute most to a FR model's decision~\cite{williford2020explainable,lu2023explanation,corrRISE2024towards,mery2022black,mery2022true,Knoche_2023_WACV,Rocha_2025_ICCVW,lu2024featureguided}, while textual explanations provide natural-language descriptions of similarities and differences between face pairs~\cite{deandres2024pixels}. Recent advances in MLLMs have enabled explanation generation for face recognition tasks. Early studies demonstrated that models such as GPT-4 can perform face verification, soft biometric estimation, and explanation generation~\cite{deandres2024chatgptface,hassanpour2024chatgpt}. Benchmarking efforts such as FaceXBench~\cite{narayan2025facexbench} evaluated MLLM face understanding capabilities across diverse tasks and prompting strategies. Domain-specific models, including FaceLLM~\cite{shahreza2025facellm} and FaceInsight~\cite{li2025faceinsight}, improve facial reasoning by incorporating specialized supervision and multimodal alignment. Prior work has also explored using foundation models for the explanation of challenging face recognition cases~\cite{sony2025foundation}. However, recent studies show that MLLMs frequently hallucinate unsupported facial attributes when generating explanations~\cite{liu2024survey,bai2024hallucination,dang2025survey,huang2025survey,zhang2025siren}. Face-centric MLLMs such as FaceInsight and FaceLLM report that explanations may rely on linguistic priors rather than visual evidence~\cite{li2025faceinsight,shahreza2025facellm}. Benchmarking studies further demonstrate that MLLMs often produce plausible but incorrect explanations in unconstrained scenarios~\cite{narayan2025facexbench}. These findings highlight the need for principled evaluation methods to assess the reliability and evidential strength of MLLM-generated textual explanations.

\section{Proposed Method}
We propose a likelihood–ratio (LR) framework for evaluating the utility of textual explanations and a multi-level prompting strategy to analyze explanation reliability.
\subsection{\textbf{Likelihood Ratio Estimation (Training Phase)}}
\label{sec:methodology-lr-train}

\noindent \textbf{\textit{i) Text Generation: }} We model the distributions of MLLM-generated 
explanations under genuine and impostor conditions. For each face pair, an MLLM 
(e.g., GPT-4o) receives the images and the ground-truth label (genuine or impostor) and produces a textual  explanation describing similarities, dissimilarities, and the verification outcome. 
We denote by $\mathcal{T}_0 \subset \mathbb{T}$ the set of explanations for genuine 
pairs and by $\mathcal{T}_1 \subset \mathbb{T}$ those for impostor pairs, where 
$T_i \in \mathcal{T}_0 \cup \mathcal{T}_1$ indexes an individual explanation sample. All models are queried via their respective public APIs with temperature $= 0.7$; 
full prompt templates and implementation details are available in the code repository.

\noindent\textbf{\textit{ii) Text Embedding and Dimensionality Reduction:}}
Each explanation $T_i$ is encoded into a fixed-dimensional vector using a frozen 
text embedding model. We use the function $f_{\text{LLM}}$, instantiated with the 
\texttt{text-embedding-3-small} model, to obtain $\mathbf{x}_i = f_{\text{LLM}}(T_i)$ where, $\mathbf{x}_i \in \mathbb{R}^d $ and $d = 1536$. We then apply principal component analysis (PCA) to reduce dimensionality while retaining 97\% of the variance, i.e., $\mathbf{z}_i = \mathrm{PCA}(\mathbf{x}_i,\ \text{var}=0.97)$, where $\mathbf{z}_i \in \mathbb{R}^k$ and $k < d$.

\noindent\textbf{\textit{iii) Gaussian Mixture Modeling:}}
In the reduced embedding space, we model the class-conditional distributions of the encoded explanations using Gaussian Mixture Models (GMMs). Separate GMMs are learned for genuine and impostor explanations: $\mathcal{P}_0(z) = \sum_{j=1}^{J_0} \pi_j^{(0)} \, \mathcal{N}\big(z \mid \mu_j^{(0)}, \Sigma_j^{(0)}\big)$ and $\mathcal{P}_1(z) = \sum_{j=1}^{J_1} \pi_j^{(1)} \, \mathcal{N}\big(z \mid \mu_j^{(1)}, \Sigma_j^{(1)}\big)$, where $\pi_j^{(\cdot)}$, $\mu_j^{(\cdot)}$, and $\Sigma_j^{(\cdot)}$ denote the mixture weights, component means, and covariance matrices, respectively, and $J_0$ and $J_1$ represent the number of mixture components under the genuine ($H_0$) and impostor ($H_1$) hypotheses. GMMs are fitted with four components and a full covariance matrix. We observed that without PCA, a diagonal covariance matrix performs better; however, after PCA dimensionality reduction, the full covariance matrix yields superior separation, as PCA decorrelates the features, making full covariance estimation more reliable and allowing the GMM to capture the true cluster geometry better.

\subsection{\textbf{Likelihood Ratio Estimation (Testing Phase)}}

At test time, a face pair $(I_A, I_B)$ is provided to the MLLM without ground-truth labels to generate a textual explanation $T$. Depending on the prompting strategy, auxiliary information from face recognition systems may also be included. The explanation is encoded using the same text embedding model and projected via the learned PCA transform to obtain the latent representation $\mathbf{z} \in \mathbb{R}^k$. The likelihood of the explanation embedding under the genuine and impostor models is then evaluated using the learned GMMs: $P_0(z) = \mathcal{P}_0(z), \qquad P_1(z) = \mathcal{P}_1(z)$ where, $P_0(z)$ and $P_1(z)$ represent the likelihood of the explanation under the genuine ($H_0$) and impostor ($H_1$) hypotheses, respectively.

The likelihood ratio and log likelihood ratio are then calculated as $
\Lambda(z) = \frac{\mathcal{P}_0(z)}{\mathcal{P}_1(z)}$ and $\log \Lambda(z) = \log \mathcal{P}_0(z) - \log \mathcal{P}_1(z)$.  Finally, we map the likelihood ratio to a bounded match score, $S_{\text{expl}}(I_A, I_B) = \frac{\Lambda(z)}{1 + \Lambda(z)}$ which provides a normalized measure of explanation strength and allows direct comparison with face recognition similarity scores \(s_{\text{FR}}(I_A, I_B)\). It is worth noting that the proposed LR framework measures the evidential consistency 
of textual explanations in embedding space rather than direct visual grounding; it 
should therefore be interpreted as a proxy measure of explanation reliability rather 
than a ground-truth faithfulness metric.

\subsection{\textbf{Multi-level Prompting}}
\label{sec:multilevel-prompting}
We adopt a multi-level prompting strategy to evaluate explanation reliability under varying levels of auxiliary FR information. Ground-truth labels are withheld during testing, and prompts provide progressively increasing information from FR models.

\noindent\textit{\textbf{i) Grounded Prompting:}} The ground-truth label (genuine or impostor) is provided with the face pair, and the MLLM generates explanations describing similarities, dissimilarities, and the verification decision (match or non-match). Used only during training data generation.

\noindent\textit{\textbf{ii) No-score Prompting:}} Only the face image pair is provided; the MLLM generates explanations based solely on visual evidence.

\noindent\textit{\textbf{iii) Score-only Prompting:}} The MLLM receives the face image pair and a similarity score from an FR system without the corresponding decision.

\noindent\textit{\textbf{iv) Score+Decision Prompting:}} The MLLM receives the face image pair, and the similarity score and the binary decision at a $0.01\%$ false match rate threshold.

GMMs are trained on grounded (genuine or impostor) explanations, while test-time labels are withheld. Since explanations are encoded via a frozen text embedding model capturing semantic meaning, the impact of this distribution shift on the GMM is expected to be minimal.

\section{Dataset Description}
\label{sec:dataset_description}
\noindent\textbf{Training:}
A subset of the BUPT-CBFace dataset~\cite{zhang2020buptcbfaceclass} containing $13{,}200$ balanced genuine and impostor image pairs is used for training. Each pair is annotated with its ground-truth label and a textual explanation generated by an MLLM. These explanations are embedded and used to train two class-conditional GMMs. A small number (23) of uncertain or undecided explanations are removed, as they do not cleanly belong to either hypothesis and would corrupt the learned distributions. At test time, however, uncertain outputs are retained and explicitly analyzed, as they reflect model confidence under challenging conditions such as extreme pose variation.

\noindent\textbf{Testing:}
We evaluate on $10{,}000$ divided equally between genuine and impostor image pairs from the IJB-S~\cite{kalka2018ijbs} dataset following the Still-to-Still verification protocol, providing varying levels of auxiliary FR information through prompting as described in Section~\ref{sec:multilevel-prompting}.

\begin{figure}
    \centering
    \includegraphics[width=0.90\linewidth]{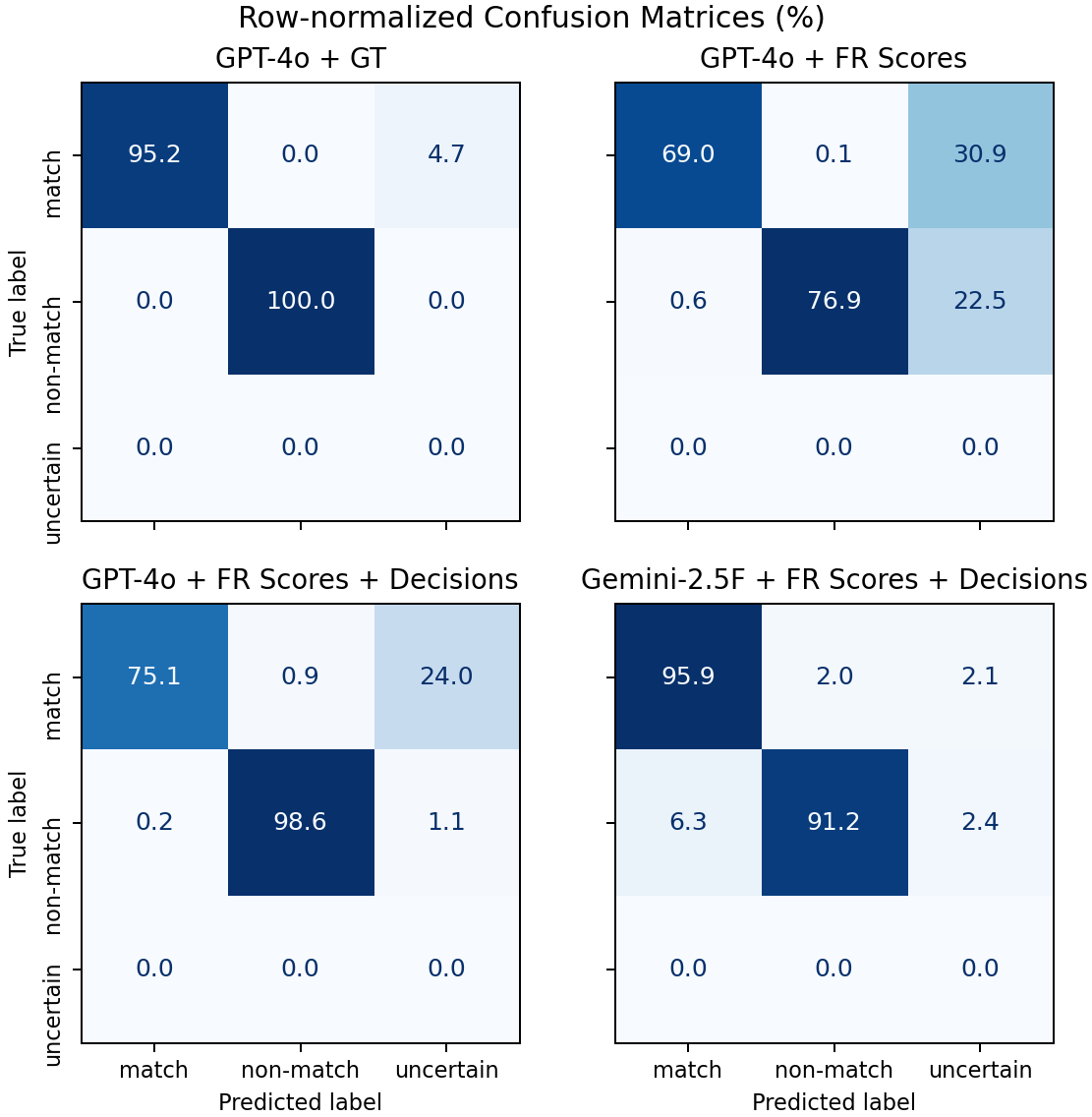}
    \caption{Row-normalized confusion matrices showing the impact of auxiliary FR information on MLLM verification decisions and uncertainty handling.}
    \label{fig:confusion_matrix}
\end{figure}

\section{Experimental Results}
We evaluate MLLMs by jointly analyzing their categorical face verification decisions and the quality of the corresponding textual explanations.  

\subsection{\textbf{Categorical Verification Analysis}}
\label{sec:mllm_classificaiton}

\begin{figure}
    \centering
    \includegraphics[width=0.98\linewidth]{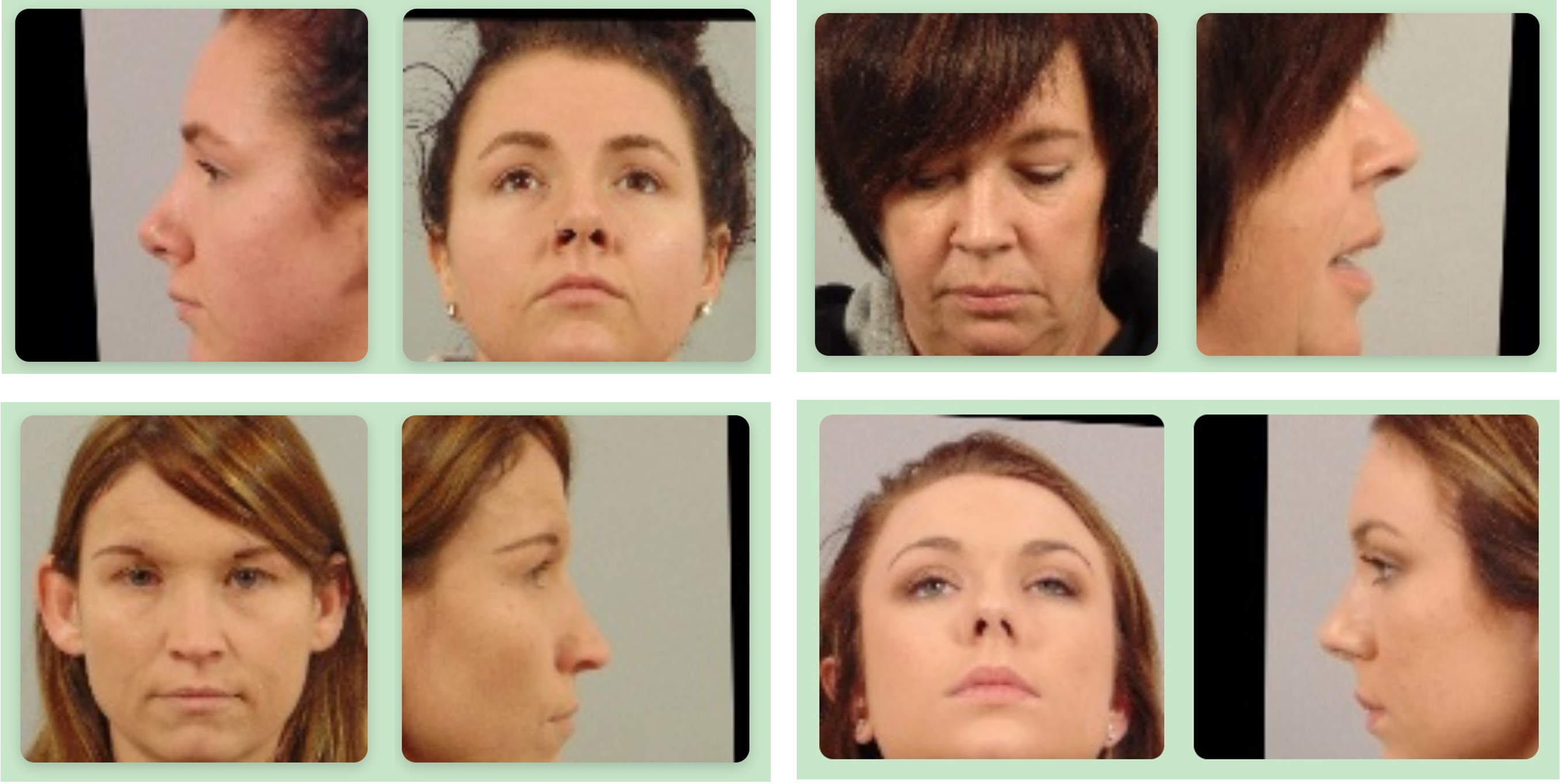}
    \caption{Examples of genuine pairs predicted as uncertain by GPT-4o despite ground-truth supervision, illustrating failures under extreme pose variation.}
    \label{fig:gpt4o_misclassified_with_gt}
\end{figure}

We evaluate verification performance based on MLLM verdicts. MLLMs output \emph{match}, \emph{non-match}, or \emph{uncertain}, where the \emph{uncertain} does not exist in the ground truth. As shown in the row-normalized confusion matrices in Figure~\ref{fig:confusion_matrix} under different prompting conditions (Section~\ref{sec:multilevel-prompting}), even when ground-truth labels are provided, GPT-4o makes errors, particularly for genuine pairs, largely due to extreme pose variations (Fig.~\ref{fig:gpt4o_misclassified_with_gt}). Providing GPT-4o with FR similarity scores improves impostor detection to $76.9\%$, but the true match rate remains low at $69.0\%$. Including both FR scores and thresholded decisions further improves impostor classification to $98.6\%$, while genuine accuracy increases modestly to $75.1\%$, with many genuine pairs still labeled as uncertain. Gemini-2.5-Flash performs better when provided with FR scores and decisions, achieving $95.9\%$ accuracy on genuine pairs and $91.2\%$ on impostors, but misclassifies $6.1\%$ of impostors as genuine. Overall, FR guidance improves impostor detection across models, but genuine verification remains challenging.

\begin{figure*}[t]
\centering

\subfloat[Dataset: CBFACE~\cite{zhang2020buptcbfaceclass}. MLLM: GPT-4o. Prompt: Ground-truth–conditioned.]
{\includegraphics[width=0.24\textwidth]{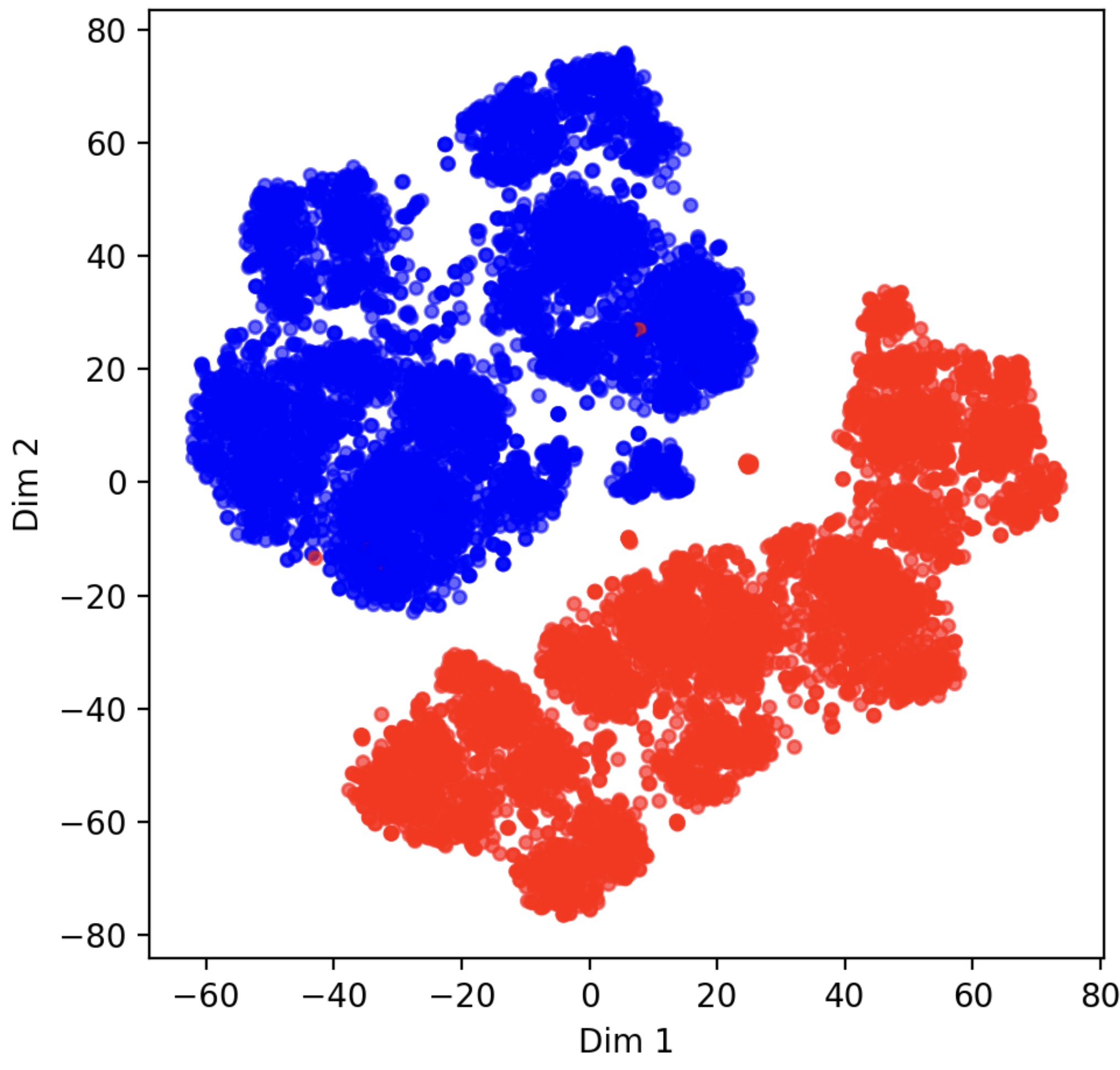}
\label{subfig:tsne_cbface}}
\hfill
\subfloat[Dataset: IJB-S~\cite{kalka2018ijbs}. MLLM: GPT-4o. Prompt: Ground truth conditioned.]
{\includegraphics[width=0.24\textwidth]{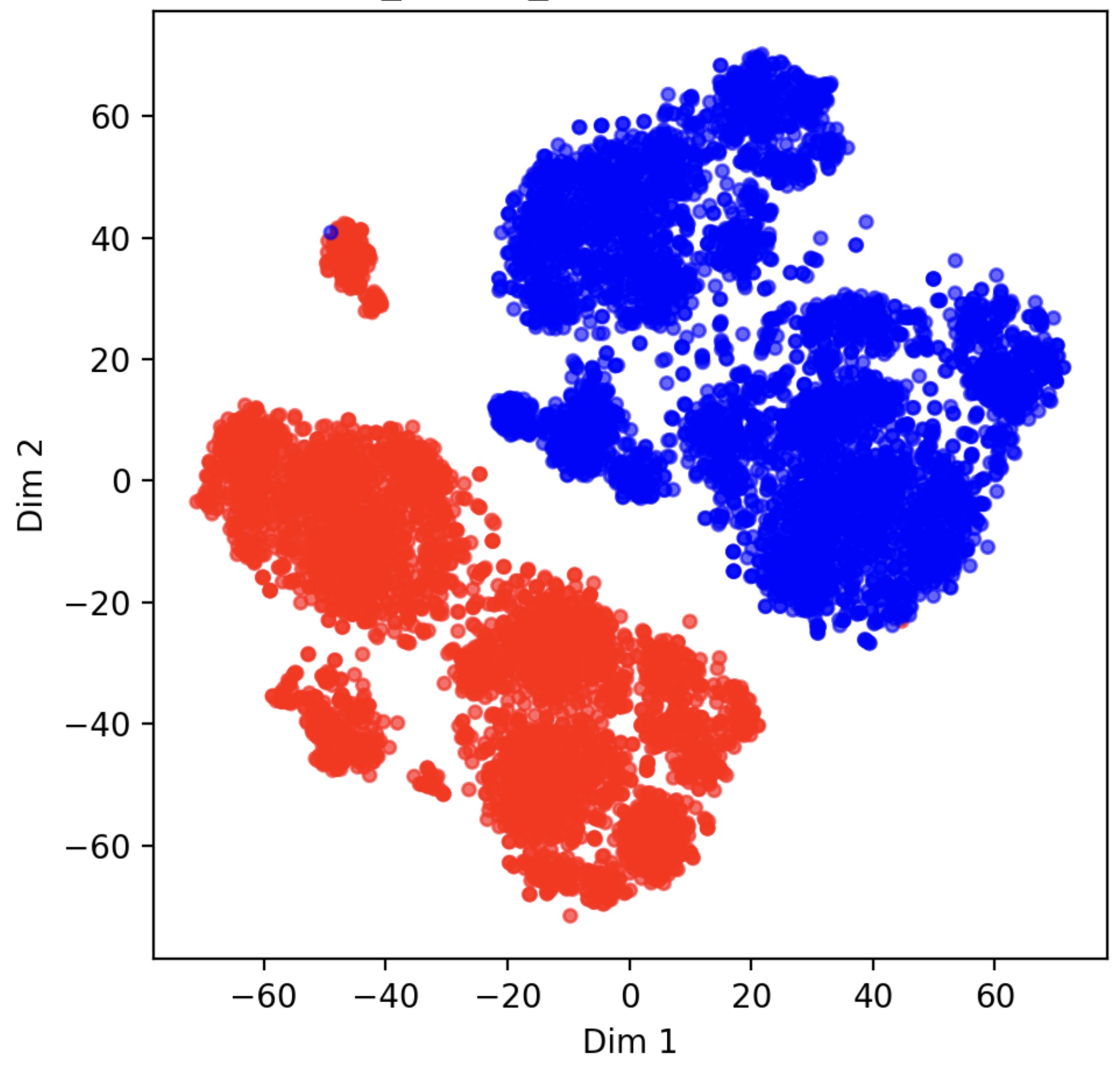}
 \label{subfig:tsne_with_ground_truth}}
\hfill
\subfloat[Dataset: IJB-S~\cite{kalka2018ijbs}. MLLM: GPT-4o. Prompt: No-score prompting.]
{\includegraphics[width=0.24\textwidth]{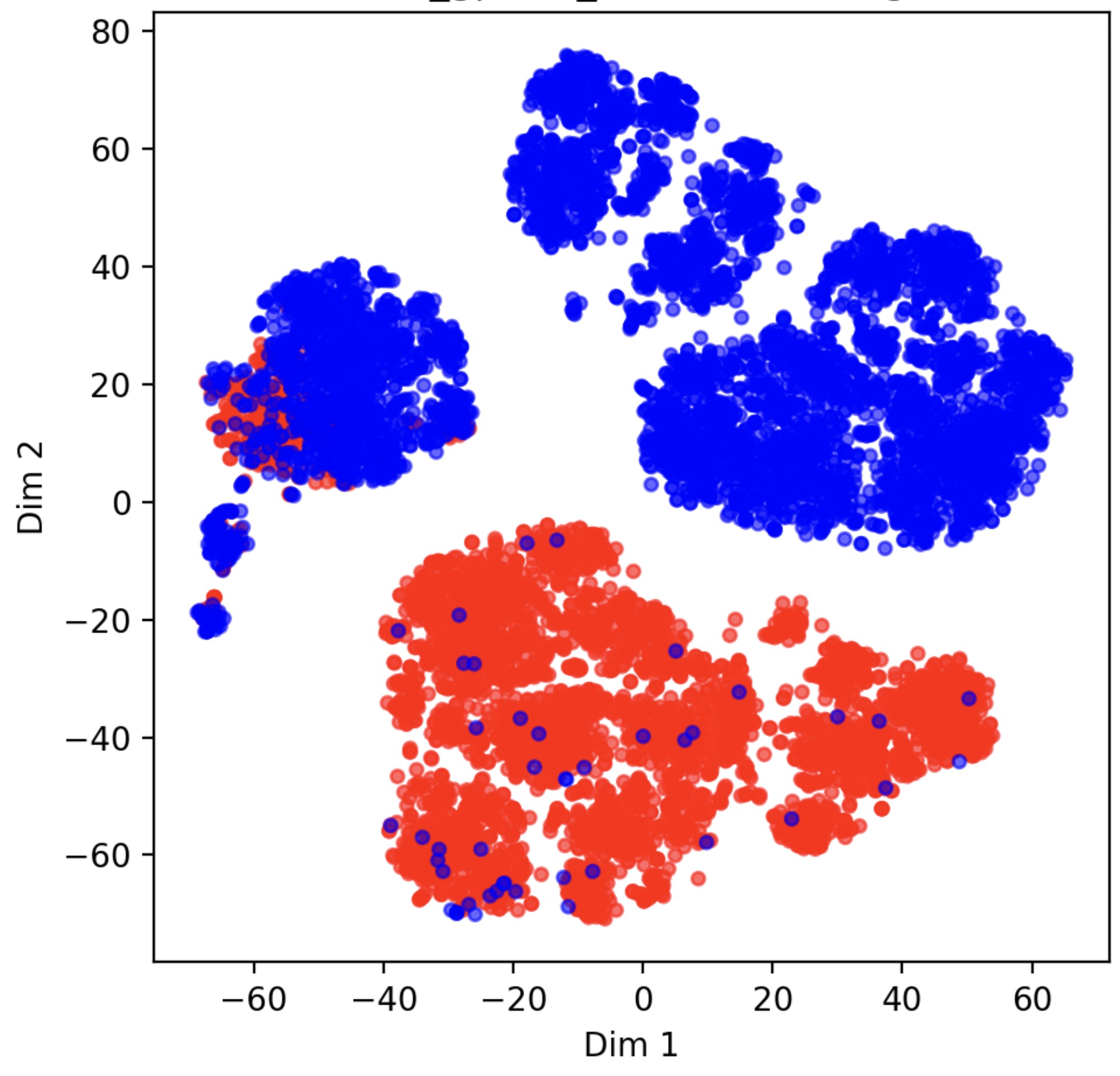}
\label{subfig:tsne_with_no_info}}
\hfill
\subfloat[Dataset: IJB-S~\cite{kalka2018ijbs}. MLLM: GPT-4o. Prompt: Conditioned on six face recognition (FR) scores.]
{\includegraphics[width=0.24\textwidth]{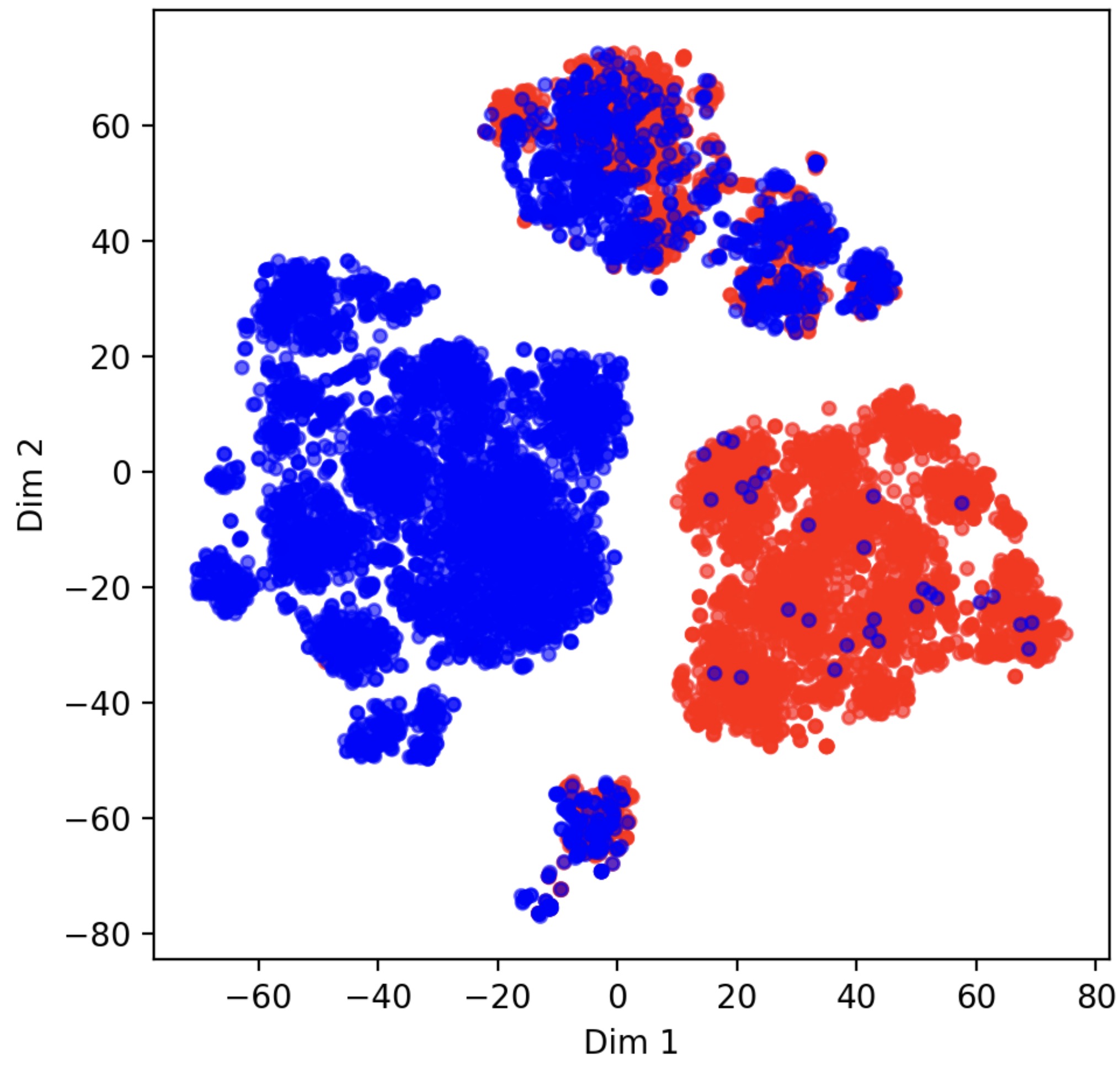}
\label{subfig:tsne_jbbs_gpt4_scores_only}}

\vspace{0.1em}

\subfloat[Dataset: IJB-S~\cite{kalka2018ijbs}. MLLM: GPT-4o. Prompt: Conditioned on six face recognition (FR) scores and hard decisions.]
{\includegraphics[width=0.32\textwidth]{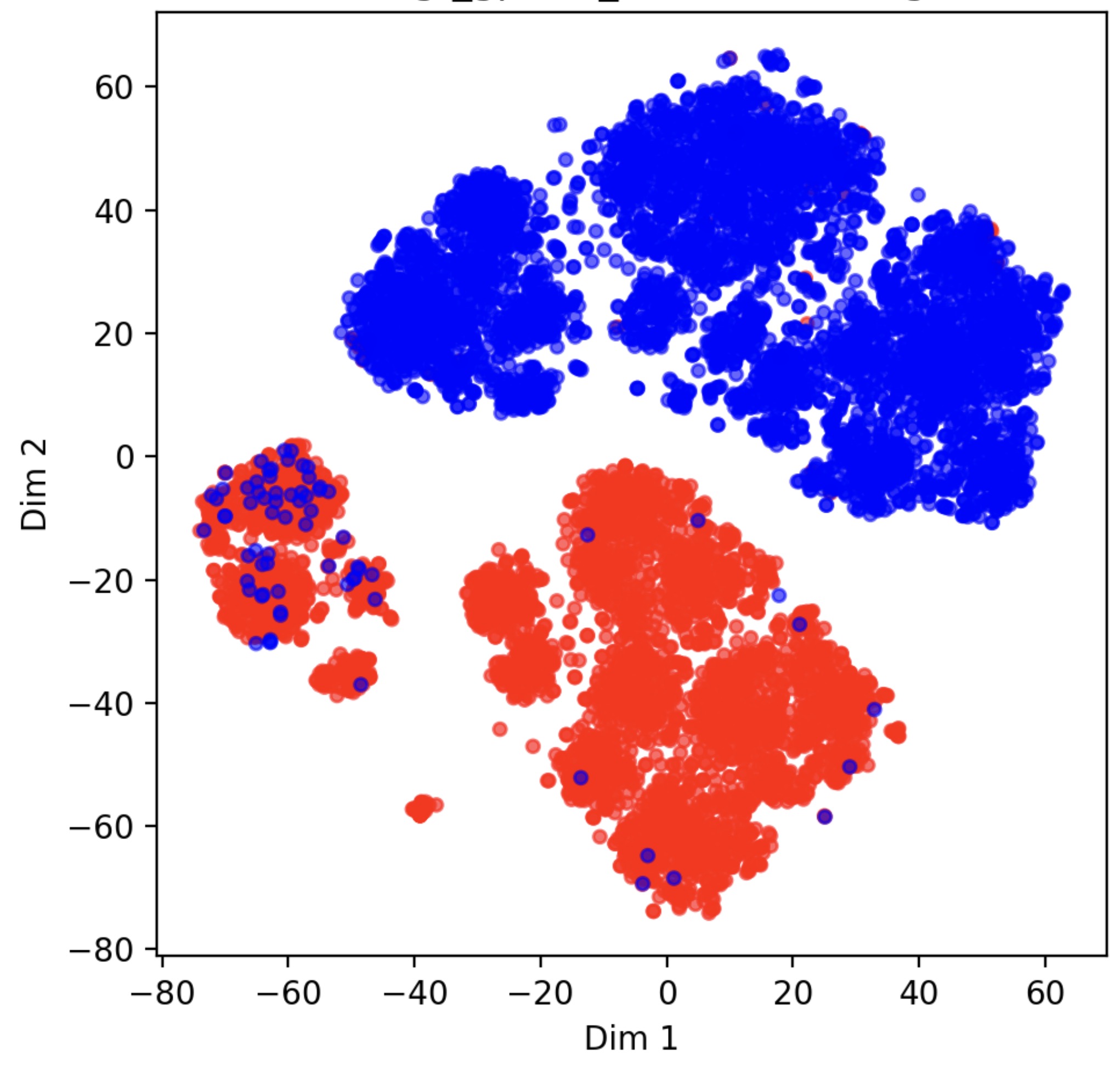}
\label{subfig:tsne_jbbs_gpt4_scores_decisions}}
\hfill
\subfloat[Dataset: IJB-S~\cite{kalka2018ijbs}. MLLM: GPT-4o. Prompt: Conditioned on KPRPE~\cite{kprpe_2022_kim} match score and hard decision.]
{\includegraphics[width=0.32\textwidth]{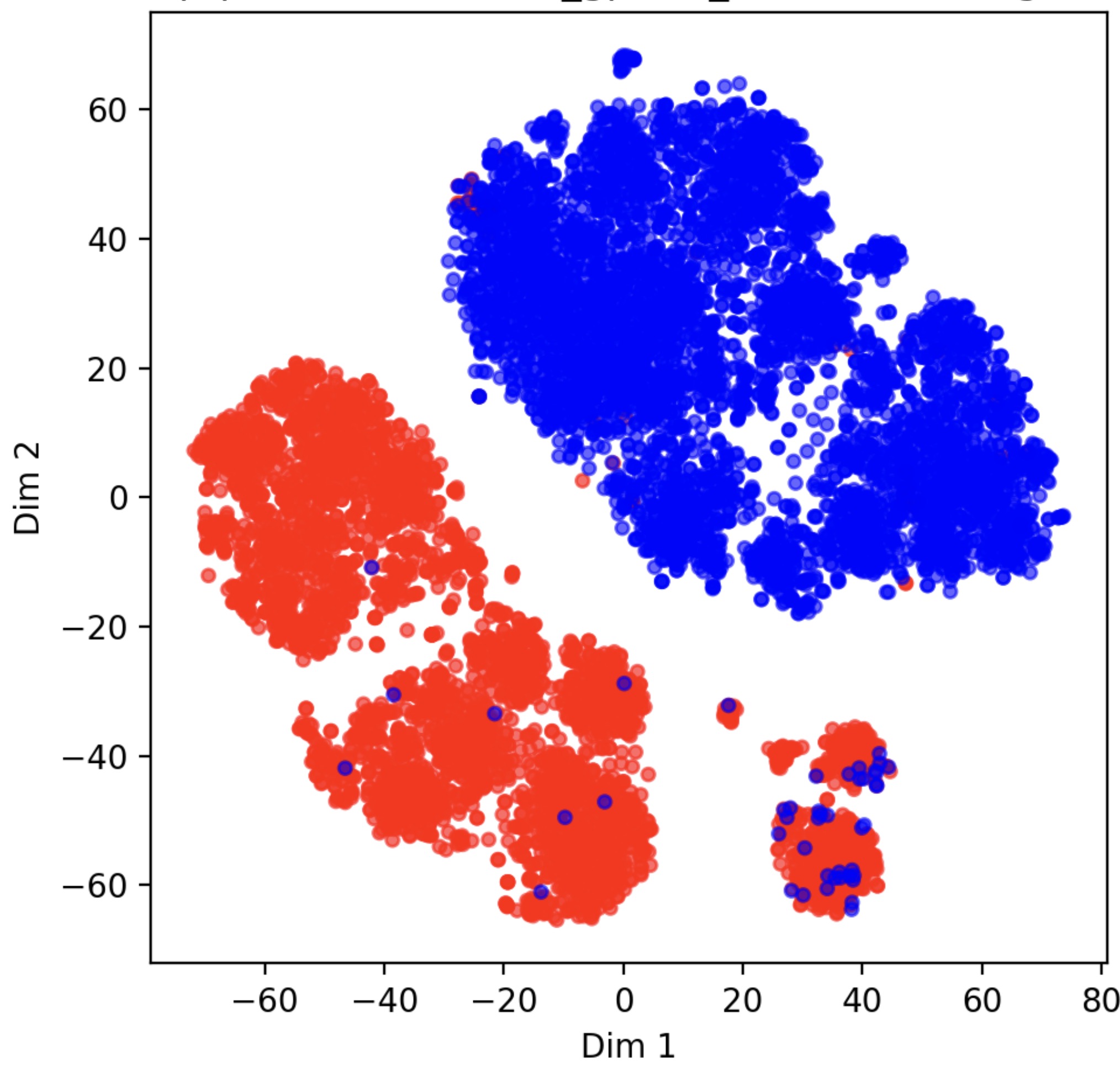}
\label{subfig:tsne_kprpe}}
\hfill
\subfloat[Dataset: IJB-S~\cite{kalka2018ijbs}. MLLM: Gemini-2.5-Flash~\cite{geminiteam2025geminifamilyhighlycapable}. Prompt: Conditioned on six face recognition (FR) scores and hard decisions.]
{\includegraphics[width=0.32\textwidth]{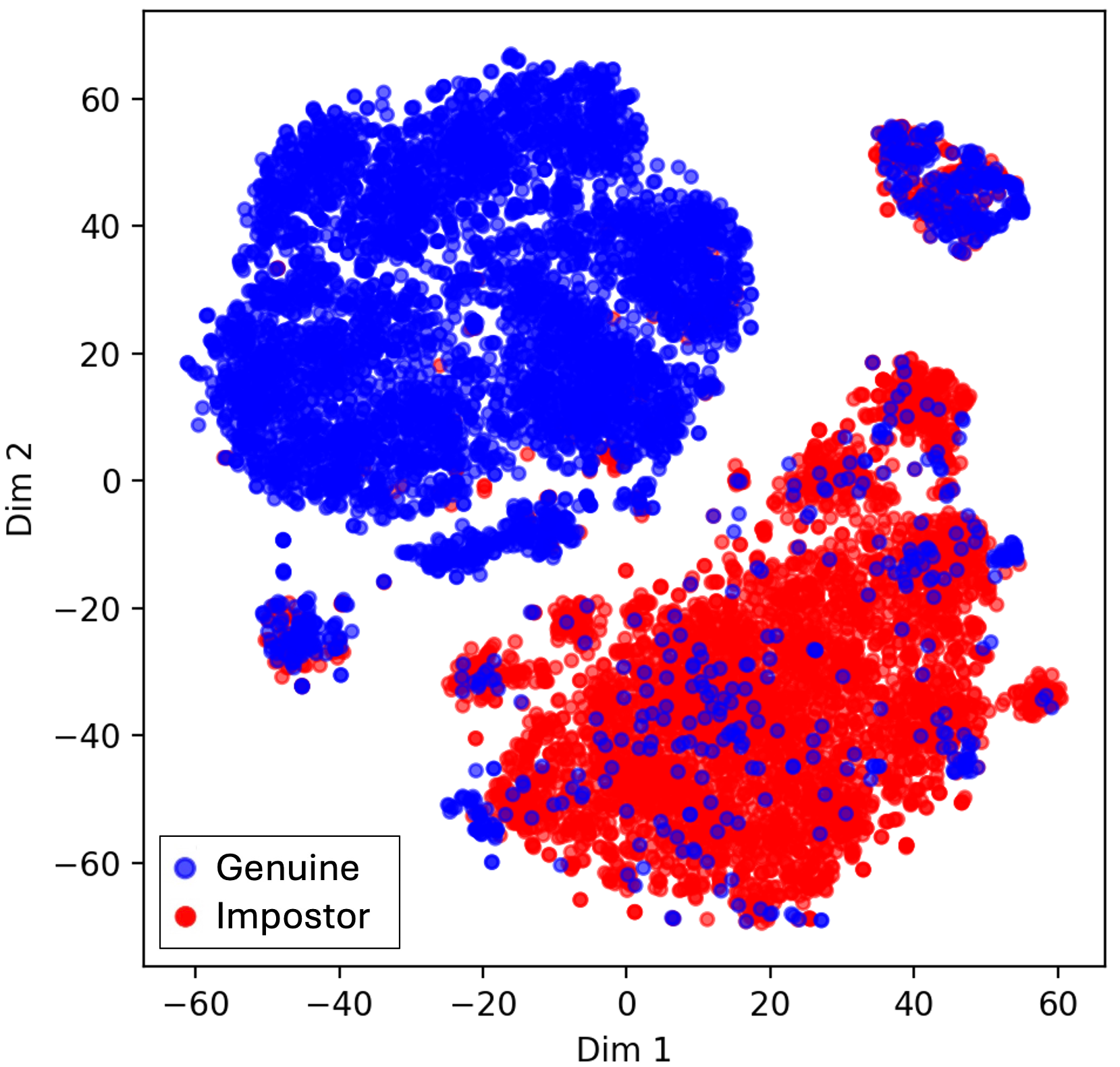}
\label{subfig:tsne_jbbs_gemini_scores_decisions}}

\caption{t-SNE visualizations of feature embeddings under different prompt strategies. The six FR model used here are ArcFace~\cite{deng2019arcface}, AdaFace~\cite{kim2022adaface}, MagFace~\cite{meng2021magface}, FaceNet-VGGFace~\cite{schroff2015facenet}, FaceNet-CasiaWebFace~\cite{schroff2015facenet}, KPRPE~\cite{kprpe_2022_kim}.}
\label{fig:tsne_all}

\end{figure*}

\begin{table}[t]
\centering
\caption{Cluster separation with metrics Silhouette coefficient, Davies-Bouldin (DB) index, Calinski-Harabasz (CH) score, inter-/intra-cluster distance ratio, and Fisher ratio are computed on the original embedding space.
$\uparrow$ indicates higher is better and $\downarrow$ indicates lower is better.
Best and second-best values are highlighted in \textbf{\textcolor{red}{red} }and \textbf{\textcolor{blue}{blue}}, respectively.}

\label{tab:cluster_separation}

\setlength{\tabcolsep}{3.5pt}
\renewcommand{\arraystretch}{1.15}
\footnotesize

\begin{tabular}{>{\centering\arraybackslash}m{2.2cm}|c|c|c|c|c}
\Xhline{1.2pt}
\textbf{Prompt Type} 
& \textbf{Silh.} $\uparrow$ 
& \textbf{DB} $\downarrow$ 
& \textbf{CH} $\uparrow$ 
& \begin{tabular}[c]{@{}c@{}}
    \textbf{Inter/Intra} \\
    $\uparrow$
  \end{tabular}
& \textbf{Fisher} $\uparrow$ \\
\Xhline{1.2pt}

GPT-4o + GT 
& \textcolor{red}{\textbf{0.30}} & \textcolor{red}{\textbf{1.38}} & \textcolor{red}{\textbf{4677.57}} & \textcolor{red}{\textbf{1.43}} & \textcolor{red}{\textbf{1.00}} \\ 

GPT-4o + No Score (\ref{subfig:tsne_with_no_info}) 
& 0.22 & 1.76 & 2837.58 & 1.27 & 0.61 \\

GPT-4o + FR Scores (\ref{subfig:tsne_jbbs_gpt4_scores_only}) 
& 0.18 & 2.13 & 1936.36 & 1.21 & 0.41 \\

GPT-4o + Scores + Decisions (\ref{subfig:tsne_jbbs_gpt4_scores_decisions}) 
& 0.25 & 1.65 & 3370.05 & 1.34 & 0.71 \\

GPT-4o + KPRPE (\ref{subfig:tsne_kprpe}) 
& \textcolor{blue}{\textbf{0.28}} & \textcolor{blue}{\textbf{1.49}} & \textcolor{blue}{\textbf{4078.40}} & \textcolor{blue}{\textbf{1.40}} & \textcolor{blue}{\textbf{0.86}} \\ 

Gemini + Scores + Decisions 
& 0.24 & 1.62 & 3358.18 & 1.31 & 0.72 \\

\Xhline{1.2pt}
\end{tabular}

\end{table}

\subsection{\textbf{Commercial FR Performance}}
To contextualize the difficulty of the IJB-S dataset, we evaluated a commercial off-the-shelf (COTS) face recognition system. The COTS system achieved $100\%$ impostor rejection and $99.69\%$ genuine accuracy, demonstrating strong performance under extreme pose variation. However, as a proprietary black-box system, it produces only numerical similarity scores without textual explanations.

\subsection{\textbf{Textual Explanation Separability and Cluster Analysis}}
We analyze whether textual explanations for genuine and impostor pairs are separable in the embedding space via cluster separability analysis on MLLM-generated explanation embeddings under different prompting strategies. Explanations are generated using GPT-4o~\cite{openai2024gpt4o} and Gemini-2.5-Flash~\cite{geminiteam2025geminifamilyhighlycapable} and embedded using a pre-trained text encoder for qualitative and quantitative analysis. Figure~\ref{fig:tsne_all} shows t-SNE~\cite{maaten2008tsne} visualizations for the training dataset (BUPT-CBFace~\cite{zhang2020buptcbfaceclass}) and test dataset (IJB-S~\cite{kalka2018ijbs}) under the Still-to-Still protocol. For training data, where ground-truth labels are provided during explanation generation, embeddings form two well-separated clusters, indicating that MLLMs produce class-consistent explanations under controlled conditions. In contrast, test embeddings exhibit greater overlap due to extreme pose variation, leading to misclassifications even when ground-truth labels are provided (Fig.~\ref{subfig:tsne_with_ground_truth}).

\begin{figure}[t]
    \centering
    \includegraphics[width=0.9\linewidth]{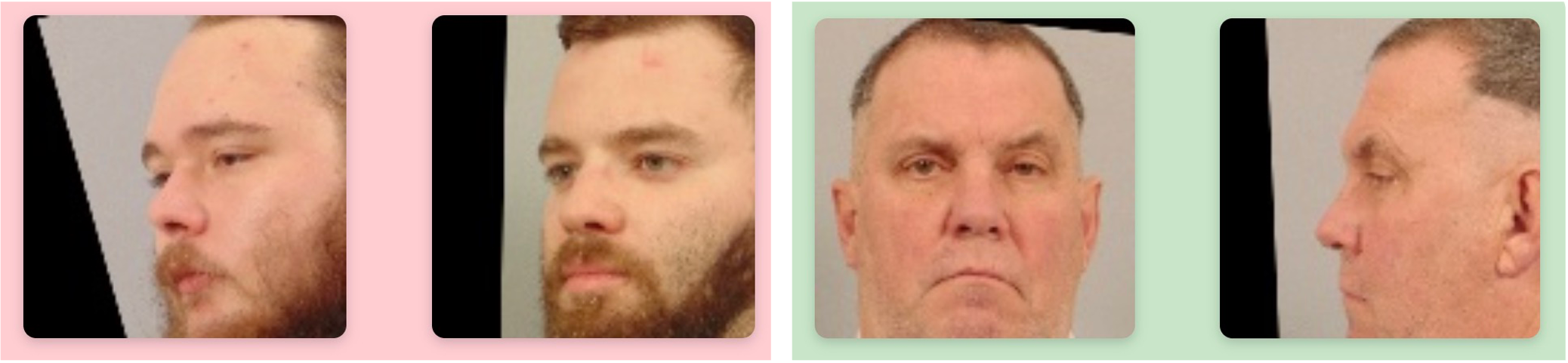}
    \caption{Examples where incorporating FR similarity scores corrected GPT-4o decisions: an impostor pair (left, pink) initially misclassified as genuine, and a genuine pair (right, green) initially marked as non-match.}
    \label{fig:improvement.}
\end{figure}

Incorporating FR match scores into the prompts improves cluster separation. Providing similarity scores (Fig.~\ref{subfig:tsne_jbbs_gpt4_scores_only}) and thresholded decisions (Fig.~\ref{subfig:tsne_jbbs_gpt4_scores_decisions}) yields clearer genuine/impostor separation, as shown in Figure~\ref{fig:improvement.} and  confirmed by quantitative metrics including the Silhouette coefficient, Davies-Bouldin index, Calinski-Harabasz score, inter-/intra-cluster distance ratio, and Fisher ratio (Table~\ref{tab:cluster_separation}). Furthermore, using a single high-performing FR model score (KPRPE~\cite{kprpe_2022_kim}) yields better separation than using multiple model scores, as multi-model inputs introduce additional variability (Fig.~\ref{subfig:tsne_kprpe}). When ground-truth labels are provided, the MLLM generates the most reliable explanations. Without them, GPT-4o struggles and frequently produces hallucinated or inconsistent outputs. While FR guidance consistently improves recognition accuracy, it does not surpass state-of-the-art face recognition performance.

\begin{figure}
\centering
{\includegraphics[width=0.48\textwidth]{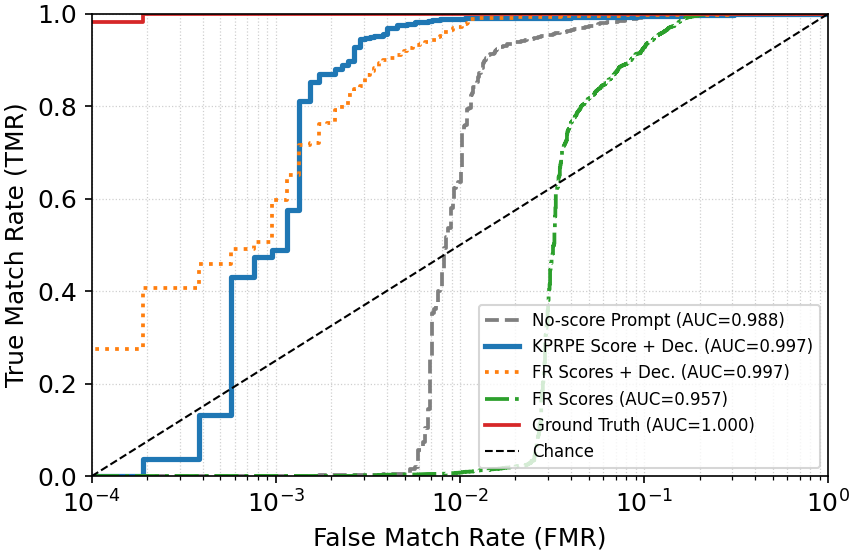}}
\caption{Likelihood-ratio–based evaluation of textual explanations. ROC curves derived from LR scores inferred from MLLM-generated explanations under different prompting strategies.}
\label{fig:lr_evaluation}
\end{figure}

\subsection{\textbf{Likelihood Ratio (LR) Evaluation}}
Figure~\ref{fig:lr_evaluation} presents the likelihood-ratio (LR)-based evaluation of textual explanations under different prompting strategies. Prompts that incorporate both FR scores and decisions, including the KPRPE (score + decision) and multi-model FR score + decision strategies, outperform the \textit{No-Score} and \textit{FR Scores Only} prompts. Among score-based approaches, KPRPE and the multi-model strategy exhibit comparable overall performance, while the multi-model prompt shows a slight advantage at lower false match rates (FMRs), suggesting benefits from aggregating information across multiple FR models.

\section{Conclusion}
While auxiliary FR information (i.e., scores and decisions) improves MLLM verification performance, it does not resolve the misalignment between textual reasoning and visual evidence, particularly under extreme pose variations. Even when MLLMs produce correct outcomes, their explanations often rely on generic or non-verifiable attributes; the correctness of the decision does not confirm the faithfulness of the explanation. This is especially critical in forensic and security applications where natural-language reasoning may serve as supporting evidence for identity decisions. Multi-level prompting improves categorical accuracy and cluster separability but does not consistently yield well-grounded explanations. We introduced a likelihood-ratio–based framework to quantify the evidential strength of textual explanations beyond categorical accuracy. A key limitation is that embedding-based separability does not directly validate visual grounding, as linguistically consistent explanations may still contain hallucinated attributes. Linking textual attributes directly to visual evidence remains an important direction for future work. However, the proposed framework is model-agnostic and can adapt to different MLLMs and prompting strategies, making it applicable as models continue to evolve.

\bibliographystyle{IEEEtran}
\bibliography{bibliography}

\end{document}